\documentclass[twoside]{article}

\usepackage[accepted]{aistats2025}

\usepackage{wrapfig}
\usepackage{amsmath}
\usepackage{amssymb}
\usepackage{mathtools}
\usepackage{amsthm}
\usepackage{float}
\usepackage{graphicx}
\usepackage[utf8]{inputenc} % allow utf-8 input
\usepackage[T1]{fontenc}    % use 8-bit T1 fonts
\usepackage{hyperref}       % hyperlinks
\usepackage{url}            % simple URL typesetting
\usepackage{booktabs}       % professional-quality tables
\usepackage{amsfonts}       % blackboard math symbols
\usepackage{nicefrac}       % compact symbols for 1/2, etc.
\usepackage{microtype}      % microtypography
\usepackage{xcolor} 
\usepackage{breqn}

\usepackage{wrapfig}
\usepackage{algorithm}
\usepackage{algpseudocode}

\newcommand{\ity}{${\mathcal{I}}$-trustworthy}

\newcommand\norm[1]{\lVert#1\rVert}

\newtheorem{problem}{Problem}

\newcommand{\bfx}{\mathbf{x}}

\newcommand{\bfz}{\mathbf{z}}

\newcommand{\bbE}{\mathbb{E}}

\newcommand{\calX}{\mathcal{X}}
\newcommand{\calZ}{\mathcal{Z}}

\newcommand{\calC}{\mathcal{C}}

\newcommand{\whf}{\widehat{f}}

\newtheorem{theorem}{Theorem}[section]
\newtheorem{proposition}[theorem]{Proposition}
\newtheorem{lemma}[theorem]{Lemma}
\newtheorem{corollary}[theorem]{Corollary}
\newtheorem{definition}[theorem]{Definition}

\usepackage{hyperref}
\hypersetup{
    colorlinks=true,% make the links colored
    linkcolor=blue,
    citecolor=blue
}  

\usepackage[round]{natbib}

\begin{document}

\twocolumn[

\aistatstitle{$\mathcal{I}$-trustworthy Models. A framework for trustworthiness evaluation of probabilistic classifiers}

\aistatsauthor{ Ritwik Vashistha, Arya Farahi }

\aistatsaddress{The University of Texas at Austin}]

\begin{abstract}
As probabilistic models continue to permeate various facets of our society and contribute to scientific advancements, it becomes a necessity to go beyond traditional metrics such as predictive accuracy and error rates and assess their trustworthiness. Grounded in the competence-based theory of trust, this work formalizes $\mathcal{I}$-trustworthy framework -- a novel framework for assessing the trustworthiness of probabilistic classifiers for inference tasks by linking local calibration to trustworthiness. To assess $\mathcal{I}$-trustworthiness, we use the local calibration error (LCE) and develop a method of hypothesis-testing. This method utilizes a kernel-based test statistic, Kernel Local Calibration Error (KLCE), to test local calibration of a probabilistic classifier. This study provides theoretical guarantees by offering convergence bounds for an unbiased estimator of KLCE. Additionally, we present a diagnostic tool designed to identify and measure biases in cases of miscalibration. The effectiveness of the proposed test statistic is demonstrated through its application to both simulated and real-world datasets. Finally, LCE of related recalibration methods is studied, and we provide evidence of insufficiency of existing methods to achieve $\mathcal{I}$-trustworthiness.
\end{abstract}

\vspace{-3mm}
\section{Introduction}
Integration of probabilistic models into AI systems continues to permeate various facets of our society \citep{abernethy2018activeremediation,raghavan2020mitigating,farzaneh2023collaborative} and contribute to scientific advancements \citep{esteves2024copacabana,ghahramani2015probabilistic}. These systems can exhibit or exacerbate undesirable biases, leading to potentially adverse and disproportionate impacts on under-represented groups \citep{pessach2020algorithmic}. This can be linked to the absence of proper evaluation tools, the lack of proper regulatory control, and the misuse of these models, hence the need for a major investment in auditing tools and regulatory solutions. This work contributes to the former by proposing a trustworthiness evaluation framework, rooted in competence-based theories of trust \citep{baier1986trust,jones1996trust,ryan2020ai,alvarado2022kind}.

\begin{figure*}[ht]
    \centering 
    \includegraphics[width=1\textwidth]{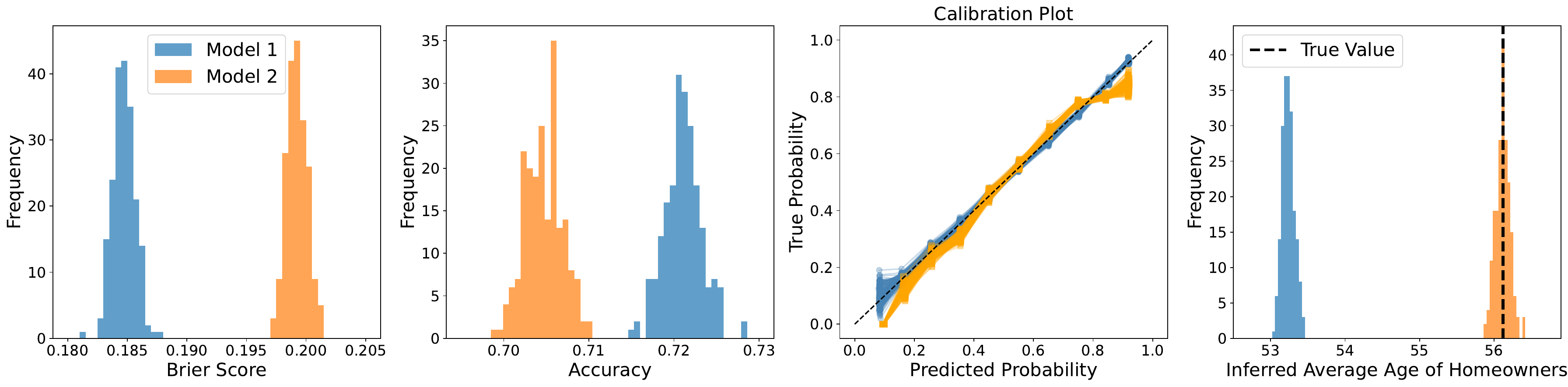} 
    \vspace{-7mm}
    \caption{\textbf{Top Panels.} Consider the task of inferring the average age of homeowners when homeownership status is unknown (test sample), using two models that assign probabilities of homeownership. While Model 1 (blue model) outperforms Model 2 in terms of accuracy, Brier score, and ECE, only Model 2 provides an unbiased estimate of the inference target (right panel). The histograms are based on 200 realizations of test/train samples. } \vspace{-3mm}
    \label{fig:limitations} 
\end{figure*} 
Predictive models are the central piece of most AI systems. A model consumes data and provides information to the next layer of a decision-making system. Majority voting and ensemble average models, which are the cornerstone of many predictive models \citep{sagi2018ensemble}, inevitably produce majority-driven biased information that might have a significant negative impact on underrepresented groups. These models produce outcomes that decrease the global rate of error, which is dominated by the majority group, instead of promoting equal results \citep{Citron:2014scored}. More importantly, historical data and human-made decisions that are used as a basis to train such models may suffer from additional systemic biases that can exacerbate the negative impacts of AI systems on underrepresented populations. There have been particular concerns around applications of predictive models to criminal justice \citep{Berk:2018fairness}, online advertising, clinical setting \citep{dreiseitl2012testing}, resource allocation \citep{farahi2024measuring}, risk assessment tools \citep{Bavitz:2018assessing}, among many other domains \citep{Citron:2014scored}. Consequently, there has been a proliferation of evaluation methods that identify and quantify the ``trustworthiness'' of the deployed models \citep{dwork2012fairness,Hardt:2016equality,Kleinberg:2017inherent,Woodworth:2017learning,Pleiss:2017fairness,Chen:2018my,Zhao:2019inherent}.  

Epistemologically, trust involves placing confidence in an information source or agent based on their ability to be accurate and reliable. In predictive models, this translates to relying on a model's consistent competence in fulfilling its intended purpose. Trust in this context follows a three-part relationship ``A trusts B to do X,'' where A is the end-user, B is the model, and X is a task \citep{horsburgh1960ethics,ryan2020ai,von2021transparency,alvarado2022kind,afroogh2023probabilistic,vashistha2024u}. Trustworthiness focuses on whether B is capable of X, leading us to rearrange our focus to ``B is trustworthy to do X.'' Following \cite{vashistha2024u}, we develop a mathematical framework to assess if ``\textbf{a predictive model} is trustworthy to perform an \textbf{inference task}.'' We argue that for a classifier to be competent in inference tasks, it must meet a stronger notion of calibration, which we term ``local calibration.''

In this work, we distinguish between two concepts: ``calibration'' and ``local calibration.'' Calibration asks whether a classifier probabilities matches the frequency of actual labels by \emph{averaging} over the entire population. In contrast, ``local calibration'' asks if a model is calibrated for every subset of the population. \cite{vaicenavicius2019evaluating} argue that a key missing element in the calibration literature is a principled approach to hypothesis testing. To fill this gap and take a step toward algorithmic trustworthiness in this work, we propose a local calibration test and a hypothesis testing method that, unlike calibration tests \citep{Widmann:2019calibration}, asks if a model is locally calibrated.

{\bf Contributions.} This work proposes a trustworthiness framework for inference tasks grounded in the competence trust theory. This includes (1) formalizing the framework, (2) specifying the sufficiency condition, (3) proposing a measure of trustworthiness, (4) developing a method for trustworthiness hypothesis testing, (5) empirically validating it, and (6) demonstrating the framework's utility in model and method evaluations.

\begin{figure*}[ht]
    \centering 
    \includegraphics[width=1\textwidth]{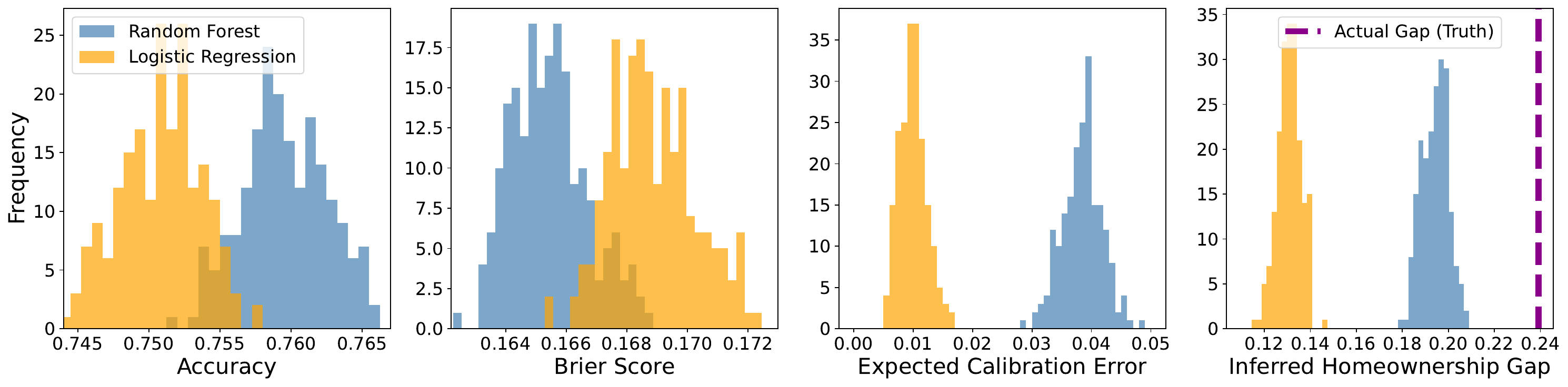}
    \vspace{-2mm}
    \caption{\textbf{Motivation 2.} Consider the task of inferring the homeownership gap between Black
and non-Black householders with two probabilistic models. Their performance is shown in the
three left panels. Both models are untrustworthy (right panel), and this lack of trustworthiness
cannot be hypothesis tested using Accuracy, Brier score, or ECE. The histograms are based on 200
realizations of test/train samples.}
    \label{fig:limitations2}
\end{figure*}

{\bf Limitations of Calibration.} 
We consider data from the 2019 American Housing Survey \citep[see Supplementary Material for a description, ][]{farahi2024analyzing} with homeownership status as the response variable for the illustrations. Suppose the aim is to estimate the age of homeowners. Figure \ref{fig:limitations} shows the result of the experiment. Model 1 outperforms Model 2 on the basis of Brier score, accuracy, and expected calibration error (ECE). However, for the task of inferring the age of homeowners, Model 2 provides an unbiased estimate, while Model 1 provides a biased estimate. In the second experiment, we consider the task of estimating the gap in homeownership between Black and non-Black households. Figure \ref{fig:limitations2} shows that the Random Forest model is better than the Logistic Regression model using accuracy, Brier score, and ECE. However, the Random Forest and Logistic Regression models fail to give an unbiased estimate of the gap in homeownership. Together, these experiments illustrate that a calibrated classifier may not be suitable for unbiased inference tasks. Additionally, these experiments challenge the perception that a calibrated classifier (Definition 2.1) is sufficient for trustworthiness in inference and the common practice that a model with better performance should be more suitable for an inference task. These observations motivate us to develop a trustworthiness framework suitable for inference.

{\bf Literature Review.} Calibration has a long-standing history in meteorology and statistics \citep{miller1962statistical, murphy1972scalar, murphy1973new, Degroot:1983comparison, Gneiting:2005weather}. Under certain assumptions, calibrating a classifier has been shown to reduce prediction bias and guarantee optimal classification performance \citep{Cohen:2004properties}. It is argued that calibration is the only requirement for trustworthy classifiers \citep{afroogh2023probabilistic, wilcox2024credences}, although this view is contested in certain decision-making contexts \citep{vashistha2024u}. The importance of calibration has led to the development of various metrics to measure and test for miscalibration \citep{kumar2018trainable, Widmann:2019calibration}. \cite{kumar2018trainable} propose a kernel-based metric and provide an optimization algorithm to obtain recalibrated estimates. Building on this work, \cite{Widmann:2019calibration} present a broader framework for measuring miscalibration using kernels. 

To address the limitations of traditional calibration, group-wise calibration -- where a model is calibrated for every identifiable group -- has been proposed as an alternative. Multicalibration, a group-wise calibration method, attempts to mitigate some of these shortcomings \citep{hebert2018multicalibration, pmlr-v209-la-cava23a}. Local Calibration is proposed as a generalized definition of group-wise calibration \citep{luo2022local, marx2024calibration,holtgen2023richness}. Recent studies propose both a measure for local calibration and a calibration method \citep{luo2022local, marx2024calibration}. This work relates Local Calibration to trustworthiness and proposes a method of hypothesis testing.

 \section{Problem Setup} \label{sec:problem_setup}

We aim to investigate the concept of trustworthiness in the context of ``B (a probabilistic model) is trustworthy to do X (an inference task).'' This work is concerned with a subset of tasks whose goal is to perform inference, hence $\mathcal{I}$-Trustworthiness. Hence, we define a $\mathcal{I}$-trustworthy model as a calibrated model (reliability) capable of achieving \emph{unbiased} inference (competence) with empirical guarantees (confidence). One can perform unbiased inference with a probabilistic classifier whose probabilities are locally calibrated on a set of covariates that have relevance to the inference task.

{\bf Notation.} We consider a probabilistic binary classification problem where $Y \in \{0, 1\}$ is the class label and $Z\in \calZ$ and $X \in \calX$ are, respectively, $d_{z}$-dimensional and $d_{x}$-dimensional instances of two, potentially overlapping, feature vectors. $Z$ is the classifier input, and $X \in \calX$ is an instance of a secondary feature vector for which we want to evaluate the model's trustworthiness. A probabilistic binary classifier, denoted with $\whf : \calZ \rightarrow \Delta$, outputs class probabilities $\whf(Z)$ of $p_{Y=1|Z}(y=1\mid z)$ for all $z \in \mathcal{Z}$. $\Delta$ denotes the probability simplex $\Delta$ := $\{\whf(z) \in \mathbb{R}_{\geq 0}: \norm{\whf(z)}_1 = 1 \}$.

\subsection{Inference Task}

The goal of an inference task is to compute the expected values of a function \( I(x) \) from inference function class $\mathcal{I}$ for the subset of a population belonging to class \( Y=1 \). Suppose the population is sampled from \( q(x) \) with \( p(y = 1 \mid X=x) \), which denotes the likelihood of each data point X = \( x \) belonging to class \( Y = 1 \) or class \( Y = 0 \). The objective is to estimate 
\begin{equation} \label{eq:inference_task}
\begin{split}
   \mathbb{E}[I(x) | y=1] =  \int  I(x) \frac{ p(y = 1 \mid x) q(x) \,}{\int p(y = 1 \mid t) q(t) \, dt}{\rm d}x,  I \in \mathcal{I}.
\end{split}
\end{equation} 
To derive an unbiased estimator of \( \mathbb{E}[I(x) \mid y=1] \), access to \( p(y = 1 \mid X=x) \) is necessary. Therefore, without a locally calibrated classifier, achieving unbiased estimation is infeasible.

\begin{definition}[{\bf Calibration}] \label{def:calibration}
$\whf$ is calibrated if $p(y = 1 \mid  \whf(z) = \alpha) = \alpha$ for all $\alpha \in \Delta$, where
$p(y = 1 \mid \whf(z) = \alpha) =\sum_{\{z, x\}  \in \calZ \otimes \calX}p(y = 1 \mid x, z, \whf(z) = \alpha)  \times  p(x \mid z,  \whf(z) = \alpha) \times p(z \mid  \whf(z) = \alpha)$.
\end{definition}
\begin{definition}[{\bf Local Calibration}] \label{def:local_calibration}
$\whf$ is locally calibrated on $\calX$ if it satisfies
$p(y = 1 \mid x, \whf(z) = \alpha) = \alpha$ for all $x \in \calX$ and $\alpha \in \Delta $, where $p(y = 1 \mid x, \whf(z) = \alpha) =  \sum_{z  \in \calZ} p(y = 1 \mid x, z, \whf(z) = \alpha) \times p(z \mid  \whf(z) = \alpha)$.
\end{definition}

Local calibration is a conditional property that implies model $\whf$ is calibrated for every subset of $\calX$; and is a stronger condition than calibration. Local calibration and multicalibration of \cite{hebert2018multicalibration} are equivalent when $\mathcal{C}$, an identifiable space as defined in \cite{hebert2018multicalibration}, and $\calX$ are equivalent and the map from $\calC$ to $\calX$ is bijective. A locally calibrated model is multicalibrated, but a multicalibrated model is not necessarily locally calibrated. Finally, we denote that if $\calZ = \calX$, then a locally calibrated model is equivalent to the Bayes optimal solution, and the condition on $\whf$ can be dropped.

Suppose $p(y = 1 \mid X = x, \whf(z)=\alpha)$ is the true frequency that a future observation belongs to class 1 conditioned on $X$ and $\whf(z)=\alpha$, where $\alpha$ is the estimated class probability via model $\whf$. Probabilistic interpretation of $\whf$ implies that if $\whf$ is calibrated then the fraction of random realizations of instances of $\{X,Z\}$ with $\whf(z)=\alpha$ where that falls in class $Y = 1$ should be $\alpha$. If the true fraction is larger (smaller) than $\alpha$ then model $\whf$ is under (over) confident in its prediction, which is undesired and the model would be miscalibrated.

$X$ consists of all features that have relevance to our inference task. It can overlap with $Z$ but they do not need to be identical. In most practical situations, $X$ is found to be low-dimensional. The difference between $X$ and $Z$ is illustrated in the following examples.

{\bf Example 1.}
Consider a different scenario where $Z$ is a vector of homeowner characteristics, and the task is to predict the likelihood of homeownership. The inference task here is to estimate the difference in average age between Black and white homeowners. Here, $X = \{$Age, Race$\}$. Local calibration examines whether the probabilities assigned to individuals of a given age and race are calibrated. Here,
$X \subset Z$. Note that $X$ overlaps with $Z$, but they are not identical

{\bf Example 2.}
Suppose $Z$ denotes CT images from patients, where the task is to predict the likelihood of breast cancer. Let the inference task be estimating the average age of Black and white patients with breast cancer. Here, $X = \{$Age, Race$\}$. Local calibration would ask whether the probabilities assigned to each patient, conditioned on their age and race, are calibrated. In this case, $Z$
is high-dimensional, while $X$ is low-dimensional, and $X \not\subset Z$. 

Note that $X$ can also contain a set of features that were not accessible or used for training. For example, a model builder did not have access to a set of protected features due to privacy issues; but an evaluator may want to guarantee the trustworthiness of the model with respect to the protected features.
\subsection{$\mathcal{I}$-Trustworthy}
Next, we seek to formalize the proposition ``B is trustworthy to do X'' for the class of population inference tasks defined in Equation~\eqref{eq:inference_task}.

\begin{definition}[{\bf \ity}]
    A model, denoted with $\widetilde{f}(.)$, is \ity\ if and only if $\widetilde{f}(.)$ is locally calibrated on $\calX$. 
\end{definition} 

{\bf Reliance.} The reliance condition is met when model $\widetilde{f}(.)$ is calibrated (Definition~\ref{def:calibration}), implying that the output of the model can be interpreted as a probability score and used as a weight in the estimator.

{\bf Competency (Theorem~\ref{th:key_theorem}).} The competency condition is met when $\widetilde{f}(.)$ is locally calibrated (Definition~\ref{def:local_calibration}). We denote that the vector $x$ depends on the inference task, and a model that is competent for one inference task might be incompetent for another inference task.  

{\bf Confidence (Collary~\ref{collary:acceptance_region}).} By hypothesis testing local calibration, the evaluator can establish statistical confidence in the trustworthiness of the model. Making a social argument to challenge the status quo requires solid empirical and statistical evidence to justify why a model is biased. We, therefore, take the position that the burden of proof is on the evaluator to demonstrate that a model is not trustworthy. To this end, the null hypothesis is that a predictive model is locally calibrated or \ity. This work equips evaluators and auditors with tools for such hypothesis testing. 

{\it Hypothesis testing setup.} Suppose $p(y \mid x, \whf(z)=\alpha)$ is unknown; we only have a ``test sample'' of size $n$ which consists of tuples of $\{x_i, y_i, \whf(z_i)\}$ for $i \in \{1, \cdots, n\}$. The test sample should be different from the sample employed for training. We further assume that they are i.i.d. samples from the joint pdf $p(x, y, z)$.  Our aim is to test the null hypothesis: ``model $\whf$ is locally calibrated on $\calX$.''  While the present work is only concerned about a binary classifier, an extension of our model to a multi-class classifier is straightforward.

\section{A Local Calibration Test Statistic and its Properties}

In this section, we propose a kernel-based local calibration test statistic and a procedure to perform local calibration hypothesis testing.

\subsection{Local Calibration Error (LCE)}

A hypothesis testing requires three elements: (i)~a null hypothesis, (ii)~a test statistic, and (iii)~an estimator of the test statistic in a finite sample setting with established convergence bounds. We specify the null hypothesis in Problem \ref{prob:main}, we construct a test statistic in Theorem \ref{th:key_theorem}, and propose an estimator for our test statistic in Equation \eqref{eq:kernel_estimator_unbiased}.  

\begin{problem}[{\bf Null Hypothesis}] \label{prob:main}
Let $\whf$ be a calibrated model at all probability levels $\alpha \in [0, 1]$. The null hypothesis is $p(y = 1 \mid X = x, \whf(z)=\alpha) = \alpha$, for all $x \in \calX$ and all $\whf(z) \in \Delta$  under probability measure $p(y, x, z)$.
\end{problem}

A model $\whf$ is locally calibrated if $p(y = 1 \mid \whf(z), x) - \whf(z) = 0$, almost surely for all $\whf(z) \in \Delta$ and $x \in \mathcal{X}$. We rewrite the equation and say model $\whf$ is locally calibrated if $\delta_c(\whf(z),x)=0$ almost surely for all $\whf(z) \in \Delta$ and $x \in \mathcal{X}$, where $\delta_c(\whf(z),x):=p(y=1 \mid \whf(z),x)-\whf(z)$. For ease of notation, we use $\delta_c$ in remainder of the text instead of $\delta_c(\whf(z),x)$. Additionally, it is often straightforward to calibrate a model via matching predicted probabilities and empirical frequency \citep{Platt:1999probabilistic,niculescu2005predicting,Zadrozny:2002transforming,Guo:2017calibration,Naeini:2015obtaining}. We, therefore, assume calibration is satisfied in the rest of this work. 

\begin{definition}({\bf Local Calibration Error}).  Let $\mathcal{H}$ be a class of functions $h: \Delta .
\rightarrow \mathbb{R}$. Let $\mathcal{G}$ be a class of functions $g: \mathcal{X} \rightarrow \mathbb{R}$. Then the local calibration error $\mathrm{LCE}$ of model $\whf$ with respect to class $\mathcal{H}$ and class $\mathcal{G}$ is
{\small
\begin{align}\label{eq: lce}
 \mathrm{LCE}[\mathcal{H}, \mathcal{G}, \whf] &:= \sup _{h \in \mathcal{H}, g \in \mathcal{G}} \mathbb{E}[\delta_{c}(\whf(Z),X) \times h(\whf(Z))g(X)] \nonumber \\ &=  \sup _{h \in \mathcal{H}, g \in \mathcal{G}} \mathbb{E}\left[\langle \delta_c, h(\whf(Z))g(X)\rangle_{\mathbb{R}}\right],
\end{align}}
where $\mathcal{H}$ is a unit ball in a RKHS $\mathcal{F}$ and $\mathcal{G}$ is a unit ball in a RKHS $\mathcal{K}$.
\end{definition}

Here $\mathcal{H}$ is the class of functions $h: \Delta \to \mathbb{R}$, where $\Delta$ is the probability simplex.  $\mathcal{G}$ is the class of functions $g: \mathcal{X} \to \mathbb{R}$, where $\mathcal{X}$ represents the feature space on which local calibration is evaluated. The function classes $\mathcal{H}$ and $\mathcal{G}$ can be impractical to work with in finite sample settings; therefore, we define $\mathcal{H}$ and $\mathcal{G}$ to be a unit ball in RKHS $\mathcal{F}$ and $\mathcal{K}$ respectively.
\begin{lemma}(cf. Lemma \ref{lemma-lceis0})  \label{remark:local_calibration}
 $\mathrm{LCE}[\mathcal{H}, \mathcal{G}, \whf] = 0$ if and only if model $\whf$ is locally calibrated.   
\end{lemma}
Now, if we set $h(.) = 1$, we get the calibration error  \citep[CE,][]{Widmann:2019calibration}, which measures calibration. We further reformulate LCE to connect it to CE so that we can use results from \cite{Widmann:2019calibration}. 

Consider the outer product  $\mathcal{M}$ = $\mathcal{H} \otimes \mathcal{G}$ where for each $h \in \mathcal{H}$ and $g \in \mathcal{G}$, we define a function $h \otimes g$ such that $(h \otimes g)(\whf(Z),X) = h(\whf(Z))g(X)$. This forms a new class of functions which belong to the RKHS $ \mathcal{P} = \mathcal{F} \otimes \mathcal{K}$.

Let $k: \Delta \times \Delta \rightarrow \mathbb{R}$ be the kernel of RKHS $\mathcal{F}$ and $l: \mathcal{X} \times \mathcal{X} \rightarrow \mathbb{R}$ be the kernel of RKHS $\mathcal{K}$. Then kernel $p$ of $\mathcal{P} = \mathcal{F} \otimes \mathcal{K}$ is tensor product of kernels $k(.,.)$ and $l(.,.)$ of $\mathcal{F}$ and $\mathcal{K}$ respectively, defined as $p((\whf(Z),X),(\whf(Z'),X')) = k(\whf(Z),\whf(Z'))l(X,X')$. Now, we can reformulate the LCE in terms of $\mathcal{M}$
$$
\mathrm{LCE}[\mathcal{H}, \mathcal{G}, \whf]:= \sup _{m \in \mathcal{M}} \mathbb{E}[\delta_{c} \times m(\whf(Z),X)],
$$
where $m(\whf(Z),X) = h(\whf(Z))g(X)$ for $h \in \mathcal{H}$ and $g \in \mathcal{G}$. However, evaluating $\mathrm{LCE}[\mathcal{H}, \mathcal{G}, \whf]$ can still be computationally intractable, thus, we come with a kernel based metric that is computationally tractable and is general enough to provide enough coverage for such hypothesis testing.
Here, the choice of $\mathcal{H}$ and $\mathcal{G}$ (and their corresponding RKHS $\mathcal{F}$ and $\mathcal{K}$) determines the sensitivity of the LCE and has an impact on Type-II error. In practice, these function classes are instantiated using kernels (e.g., RBF kernels). Different kernel choices will instantiate different function classes and corresponding RKHS. In our work, we have used the RBF kernel, but our theoretical results hold for all types of universal kernels. 
\subsection{A kernel-based Equivalent of $\mathrm{LCE}$}
We simplify $\mathrm{LCE}$ by using the reproducing property of Kernels in RKHS. We define the following kernel based equivalent of $\mathrm{LCE}$.

\begin{definition}
    Kernel $\mathrm{LCE}$ ($\mathrm{KLCE}$) with respect kernel $l(.,.)$ and $k(.,.)$ is defined as $ \mathrm{KLCE}[k, l, \whf] := \mathrm{LCE}[\mathcal{H}, \mathcal{G}, \whf]$.
\end{definition}

\begin{theorem}[{\bf \ity\ Competency Theorem}] (cf. Theorem \ref{th-formula}) \label{th:key_theorem}
Let $(X^{\prime}, Z^{\prime} ,Y^{\prime})$ be an independent copy of $(X, Z, Y)$ then
\begin{align} \label{eq:pre_key}
 &\mathrm{KLCE}[k, l, \whf]^2 \\ &=\mathbb{E}[(Y-\whf(Z))k(\whf(Z),\whf(Z'))l(X,X') (Y'-\whf(Z'))] \nonumber 
\end{align}
\end{theorem}  
$\whf$ is \ity\ if and only if ${\rm KLCE}^2[\cdot] = 0$.  % This implies that $\mathrm{KLCE}$ is a proper scoring rule.

Assuming model $\whf$ is calibrated and ${\rm KLCE}^2[\cdot] \neq 0$ implies there are biases that changes with group but cancels on average. In this case, the model would be considered untrustworthy with respect to $x$ but calibrated on average.

An unbiased, consistent estimator of Equation~\eqref{eq:pre_key}, with a computational cost of $\mathcal{O}(n^2)$, is
\begin{align} \label{eq:kernel_estimator_unbiased}
     &\widehat{{\rm KLCE}^2}[k, l, \{x, y, z\}, \whf]   \\ &= \frac{1}{n(n-1)}  \sum_{i\neq j}^{n} [ (y_i - \whf_i) \, k(\whf_i, \whf_j) \, l(x_i, x_j) (y_j - \whf_j) ] \nonumber
\end{align}
A generalization to multi-class classifiers is straightforward. Suppose we have $m$ classes, then $y$ is an $m$-dimensional unit vector that specifies class membership, and $\whf$ is a probabilistic model that for every input $z$ outputs a prediction $\whf(z) \in \Delta^{m}$ where $\Delta^{m}$ denote the $(m-1)$-dimensional probability simplex. See \cite{Widmann:2019calibration} for proofs and conditions. However, it is important to note that \ity\ focuses on one class at a time; therefore, for the purposes of \ity, binary classification is sufficient.

Setting $k(., .) = 1$ tests the calibration of model $\whf$ and we recover $\mathrm{KCE}$ of \cite{Widmann:2019calibration}. Since $\mathrm{KCE}$ is a special case of $\mathrm{KLCE}$, we do not compare them. We can also recover the measure used in \cite{marx2024calibration} for their optimization based calibration method by setting $\mathcal{Z} = \mathcal{X}$. The distinction between $\mathcal{Z}$ and $\mathcal{X}$ allows us to evaluate the model's trustworthiness on a potentially different set of features than those used for training the model. Our kernel $l(x_i, x_j)$ operates on this secondary feature space, enabling a more flexible assessment of local calibration.

\begin{corollary} \label{prop:metric}
    The squared $\mathrm{KLCE}$ of Equation \eqref{eq:pre_key} is a metric. Hence, it can be used to estimate the distance between model probability output and actual class frequency.
\end{corollary}
The $\mathrm{KLCE}$ is a metric when $\mathcal{F}$ and $\mathcal{K}$ are a universal RKHS. Since the corresponding kernels used in this work are characteristic, they are universal RKHSs; hence the Corollary. $\mathrm{KLCE}$ statistic quantifies how close the prediction of a model is to the actual class probability, thus it may be used for model comparison and model selection as well. A model with smaller $\mathrm{KLCE}$ can be considered as a less untrustworthy model.

\begin{theorem}[{\bf Convergence Bound}] (cf. Theorem 
\ref{th: test-proofs}) \label{th: test-proof}
Let $p: (\Delta, \mathcal{X}) \times (\Delta, \mathcal{X}) \rightarrow \mathbb{R}$ be a kernel, and assume that $p(\cdot, t) u$ is measurable for all $t \in \Delta$ and $u \in \mathbb{R}$, and $P_{\alpha ; \beta}:=\sup _{s, t \in \Delta}\|p(s, t)\|_{\alpha ; \beta}<\infty$ for some $1 \leq \alpha, \beta < \infty$. Then for all $\epsilon>0$
$$
\begin{aligned}
 \mathbb{P}\left[ \left|\widehat{\mathrm{KLCE}^2}[.]-\operatorname{KLCE}^2[.] \right| \geq \epsilon\right]   \leq 2\exp \left(-\frac{\epsilon^2 n}{2 P_{\alpha ; \beta}^2}\right)   
\end{aligned}
$$
\end{theorem}
 By setting $\frac{\epsilon^2 n}{2 B_{\alpha ; \beta}^2} = -\ln \alpha_p$ and solving for $t$, we get the convergence rate and following corollary.

\begin{corollary}[{\bf \ity\ Confidence Gaurantee}]
    \label{collary:acceptance_region}
A hypothesis test of level $\alpha_p$ for the null hypothesis ${\rm KLCE} = 0$ has the acceptance region
\begin{equation}
    \widehat{{\rm KLCE}^2}[k, l, \{x, y, z\}, \whf] < \frac{B_{\alpha ; \beta}}{\sqrt{n}}\sqrt{\ln \alpha_p^{-2}}. 
\end{equation}
\end{corollary}

We conclude that Equation \eqref{eq:kernel_estimator_unbiased} is a consistent estimator of $\mathrm{KLCE}$ and converges at a rate of $\sqrt{n}$. 
% Given a hypothesis testing two approximately locally calibrated models has ELCE estimate less than $\frac{\sqrt{8}K}{\sqrt{n}}\sqrt{\alpha^{-1}}$. %This implies that $\whf_{c_1}$ and $\whf_{c_2}$ can be consistent with the null hypothesis but $\whf_{c_2} \neq \whf_{c_1}$. Indeed, one can argue that an approximate classifier is always biased and only in a limit of $n \rightarrow \infty$ the true classifier can be achieved.  
%\subsection{Statistical significance in a finite sample setting}
% In a finite sample setting, we must quantify the statistical significance of a non-zero ELCE.
The statistical significance ($p$-value) is defined as the probability of the null distribution exceeding the $\mathrm{KLCE}$ for the proposed model. The $p$-value can be computed as:  
\begin{equation} \label{eq:p-value-computation}
    p = {\rm Pr}\left(\widehat{{\rm KLCE}^2}_{\rm null}[\cdots] > \widehat{{\rm KLCE}^2}_{\rm data}[\cdots]\right).
\end{equation}
Here, $\widehat{{\rm KLCE}^2}_{\rm null}$ is the estimated value of KLCE statistic under the null distribution, which is that the probabilities are locally calibrated. $\widehat{{\rm KLCE}^2}_{\rm data}$ is the estimated KLCE statistic using the data. We compute both using a simulation-based bootstrap algorithm described in detail in Supplementary Material.

\subsection{Model Criticism as a Diagnostic Tool} \label{sec:diagnosis}

Next, we develop a diagnostic strategy that guides the user to localize where and for which groups a particular model may fail to explain the observed data. We address this via a ``model criticism'' framework. The model criticism framework was developed to evaluate fitted Bayesian models and discrepancies between two data distributions \citep{marshall2003approximate,Kim:2016examples,nott2018approximation}. Motivated by statistical model criticism proposed by \cite{Lloyd:2015statistical}, we propose a principled diagnostic tool that localizes and quantifies calibration biases as a function of $x$. It is based on functions $g(.)$ and $h(.)$ which attain the supremum of Equation~\eqref{eq: lce}.

\begin{proposition} \label{prop:witness_function}
Suppose local calibration bias is defined as ${\rm LCB}(x^{\prime}) := P(y \mid X=x^{\prime}, \whf=\alpha ) - \alpha$. The supremum function, referred to as ``error witness function'', has a closed form solution \citep{Gretton:2012kernel}. The solution yields
\begin{align}
    {\rm LCB}[k, l, \{x^{\prime}, y^{\prime}, \whf(z^{\prime})\}] := \bbE \left[ \delta_c k(\whf(z^{\prime}), \whf(z))\,l(x, x^{\prime}) \right]  \nonumber
\end{align}
\end{proposition}
This function measures the maximum discrepancy between the observed data and predicted probabilities as a function of $x$.
An evaluation of the above equation localizes and quantifies bias in the feature space $x$. A properly normalized estimator of ${\rm LCB}(x^{\prime})$ is
\begin{align} \label{eq:EWF_Estimator}
    &{\rm LCB}[k, l, \{x^{\prime}, y^{\prime}, \whf(z^{\prime})\}] \\ &= \frac{\sum_{i=1}^{n} \left[ (y_i - \whf(z_i) k(\whf(z_i), \whf(z^{\prime})) l(x_i, x^{\prime}) \right]}{\sum_{i=1}^{n} \left[ k(\whf(z_i), \whf(z^{\prime})) l(x_i, x^{\prime}) \right]}. \notag 
 \end{align}
 for each observation in the test sample. Thus, one can correct the model's output with this estimated bias at an individual level. Here, if we set $k(.) = 1$ and $\mathcal{Z} = \mathcal{X}$, then we get the measure of local miscalibration proposed in \cite{luo2022local}.

\section{Experiments}

\begin{figure}[ht]
    \centering 
    \includegraphics[width=0.48\textwidth]{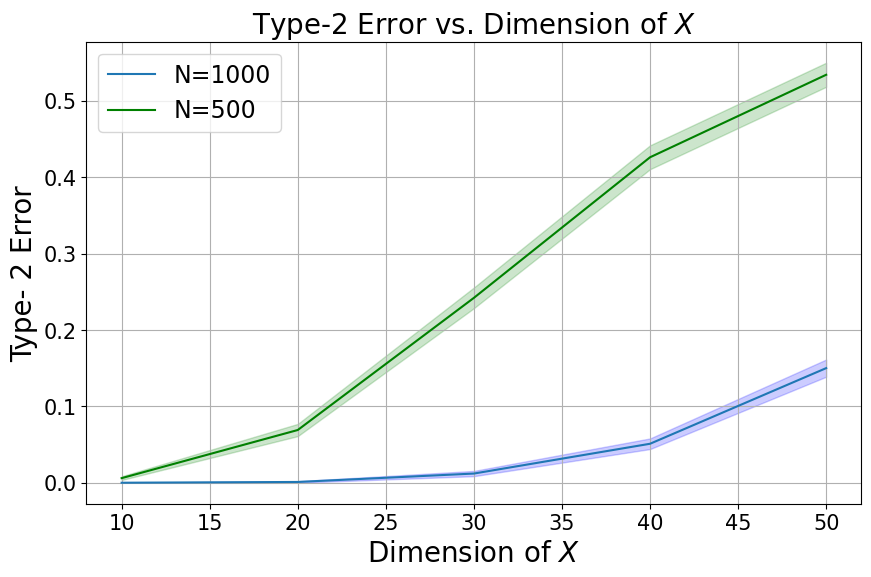}
    \vspace{-7mm}
    \caption{Type-II Error as a function of dimension of $X$ and sample size $N$.} 
    \label{fig:simulation}
\end{figure}
In this section, we discuss experiments done on both simulated and real data to illustrate the performance of our test statistic and its properties in different cases\footnote{Code for reproducing experiments is available at \hyperlink{https://github.com/ritwikvashistha/I-trustworthy}{https://github.com/ritwikvashistha/I-trustworthy}}.
{\bf Simulations.} In the simulated setting, we obtain the distribution of our test statistic and estimate the Type-II error as the dimension of the data increases. Suppose, $x_i \sim \mathcal{N}\left(0, 1\right)$ where $i \in\{1, \cdots, d\}$, and $y \sim \operatorname{Bernoulli}(\bar{p})$, where $\bar{p}=\operatorname{sigmoid}\left(\sum_{i=1}^d x_i\right)$. The Bayes classifier is $f=\operatorname{sigmoid}\left(\sum_{i=1}^d x_i\right)$. Consider the model $\widehat{f}(x)=\operatorname{sigmoid}\left(\sum_{i=1}^{d-1} x_i\right)$ where $\widehat{f}(x)$ is calibrated on $\left\{x_1, \cdots, x_{d-1}\right\}$ but miscalibrated on $\left\{x_d\right\}$. Now we hypothesis-test the calibration of model $\widehat{f}$ on $x=\left\{x_1, \cdots, x_d\right\}$ since initially one does not know along which dimension the model is miscalibrated. Figure \ref{fig:simulation} shows the Type-II error and how it varies with the dimension and sample size $N$. We generate $1000$ data realization for $N = 500$ and $N=1000$. For each data realization, we compute $\widehat{\operatorname{KLCE}^2}(.)$ for the null model and the miscalibrated model, and estimate the rate of Type-II error by counting the fraction of samples for which the test fails to reject the null hypothesis. As expected, the Type-II error increases with the data dimension. For trustworthiness purposes, we expect $X$ to be low-dimensional. Therefore, the test statistic has a good discriminative power in our operating regime. In Supplementary Material, we also look at Type-I error and how it is independent of kernel width and sample size. 

{\bf Real-world Application.} In the case of real data, we consider the COMPAS dataset \citep{Propublica2016CompasAnalysis} and show an application of our method to auditing models. The prediction task in the dataset is to predict a defendant’s possibility of re-offend. Such predictions are used to decide the bail amount, opportunity in the pretrial hearing, or even for sentencing purposes. See Supplementary Material for the description of the dataset. 

We train a binary classifier to predict the likelihood of a defendant recidivism. Our predictive model takes both the defendant's demographics (including age, gender, and race) and their criminal history. The criminal history includes: (1) crime degree, (2) a count of their prior convicted crimes, (3) the number of juvenile felony charges, and (4) counts of juvenile misdemeanor charges on their record. All these features are used to train a predictive model, but only the defendant's demographic information is used to test the model's \ity. We randomly split the data into three equal, non-overlapping subsets. We train a random forest model using the first subset; the second subset is used to train a calibration method; and the final subset to test \ity\ (local calibration on demographic features). We then employ the trained model and use the second subset (with the labels) to predict the calibration bias for defendants in the last subset.

\begin{figure*}[ht]
    \centering 
    \includegraphics[width=1\textwidth]{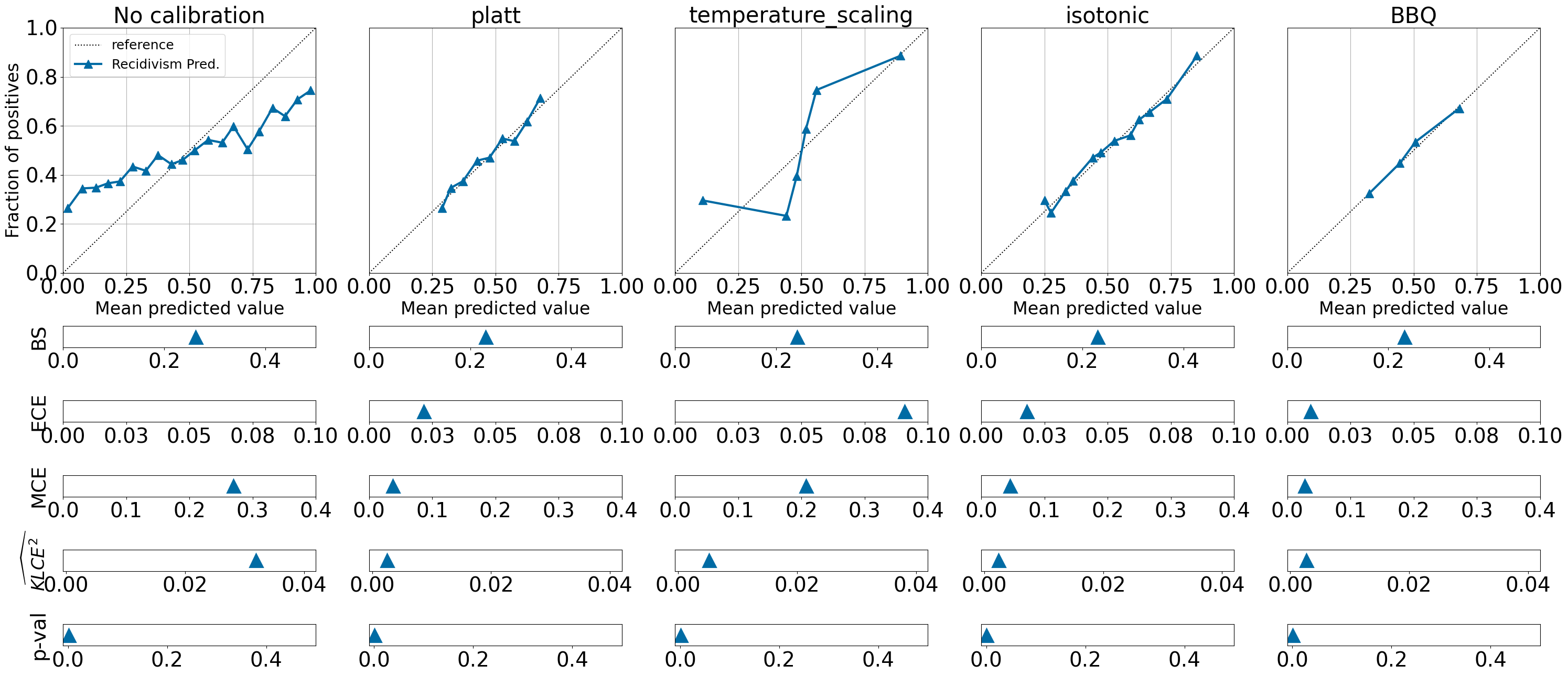} 
    \vspace{-4mm}
    \caption{Recidivism risk prediction results. Calibration error measures are shown for before calibration (left column) and after calibration. Reliability diagram (top panels) visualizes how well probabilities are calibrated. BS, ECE and MCE quantify how well a model is calibrated. ${\rm KLCE}$ test statistic can measure local calibration error. The corresponding $p$-value quantifies the statistical significance of ${\rm KLCE}^{2} = 0$. }
    \label{fig:results_test}
\end{figure*}

The leftmost panel of Figure \ref{fig:results_test} shows the calibration performance of the prediction model for recidivism risk. The results suggest that the model is locally miscalibrated. Thus, we use various calibration methods and compare there performance. Platt scaling \citep{Platt:1999probabilistic} transforms classifier outputs into probabilities using logistic regression. Temperature scaling \citep{Guo:2017calibration} adjusts outputs with a single temperature parameter to refine probability estimates without altering rankings, whereas Isotonic Regression \citep{Zadrozny:2002transforming} calibrates using a non-decreasing function to address non-linear output relationships. Bayesian Binning into Quantiles \citep[BBQ,][]{Naeini:2015obtaining}  segments predictions into quantiles and employs Bayesian updating for each segment. After post-calibration, our hypothesis testing indicates that the existing calibration methods cannot achieve local calibration. While all calibration methods reduce the magnitude of miscalibration, they are unable to obtain a locally calibrated model, supporting the fact that calibration is insufficient. This implies that the trustworthiness of this model cannot be guaranteed, and there are group-wise biases that can potentially have fairness implications. 

In Supplementary Material, we look at how the local calibration bias varies for defendants in the test sample, across gender and race.

\noindent {\bf Application in Method Evaluation.} Our method can also be used to evaluate learning methods and post-processing calibration and learning methods. To demonstrate this, we assessed various calibration methods, specifically \citep{kumar2018trainable}, multicalibration \citep{hebert2018multicalibration,pmlr-v209-la-cava23a}, and local calibration \citep{marx2024calibration}.
In our initial experiment, we obtained multicalibrated \citep{hebert2018multicalibration} and proportionally multicalibrated \citep{pmlr-v209-la-cava23a} probabilities for three models (a Random Forest, a Logistic Regression, and a constant Classifier) using the 2019 American Housing Survey data \citep{farahi2024analyzing}. The results are shown in Table \ref{tab:table 1}. 
We tested if the probabilities are locally calibrated for the variables household income and race (1 if Black, and 0 otherwise). Our findings reveal that all classifiers fail to be locally calibrated post multicalibration or proportional multicalibration. This observation supports our claim that a multicalibrated model is not necessarily locally calibrated, underscoring the importance of our measure.
We also evaluated the learning methods proposed in \cite{kumar2018trainable} and \cite{marx2024calibration}, which aim to achieve post-processing calibration and local calibration, respectively. We applied these methods to multiple datasets and compared different scores with our KLCE measure. The results are shown in Table \ref{tab:table2}. These findings demonstrate that these learning methods are not universally effective, highlighting the need for a robust test to evaluate their performance across different scenarios.
\begin{table}[H] % Table floats on the left
  \centering
  \begin{tabular}{p{1.7cm}|p{1.7cm}|p{1.2cm}|p{1.2cm}} 
    \textbf{Method} & \textbf{Constant} & \textbf{LR} & \textbf{RF} \\ \hline \hline
    None & <0.002 & <0.002 & <0.002 \\ \hline
    MC & <0.002 & <0.002 & 0.004 \\ \hline
    PMC & <0.002 & <0.002 & 0.008 \\ \hline
  \end{tabular} 
  \caption{$p$-value of local calibration with no calibration, MC (Multicalibration) and PMC (Proportional Multicalibration) for a constant, logistic regression, and random forest classifier.}
  \label{tab:table 1}
  \vspace{-3mm}
\end{table}
\begin{table*}[th!]
\centering
\begin{tabular}{lllcc}
\hline
Dataset & Training Objective & Brier Score & ECE  & $\widehat{{\rm KLCE}^{2}}$($p$-value) \\
\hline
Homeownership & XE & 0.1699 & 0.0193  & 0.0006 (0.002)  \\
$n = 48660,d = 11$ & XE + MMCE &  0.1703 & 0.0157 & 0.0001 (0.002)   \\
$X$ = Income and Race & XE + MMD  & 0.1711 & 0.0266 & 0.0003 (0.002)    \\
\hline
Adult (UCI) & XE & 0.1080 & 0.0478 & 0.0009 (0.002)  \\
$n = 32561, d = 104$ & XE + MMCE & 0.1145 & 0.0680 & 0.0014 (0.002)  \\
$X$ = Age and Gender & XE + MMD  & 0.1092 & 0.0360 & 0.0015 (0.002) \\
\hline
Heart Disease (UCI) & XE & 0.1162 & 0.1014 & 0.0031 (0.26)  \\
$n = 918, d = 12$ & XE + MMCE & 0.1130 & 0.0801 & 0.0038 (0.222)  \\
$X$ = Age and Gender & XE + MMD & 0.1150 & 0.0964 & 0.0028 (0.262) \\
\hline
\end{tabular}
\caption{Experimental results for binary classification tasks with Logistic Regression, comparing MMCE \citep{kumar2018trainable} and MMD regularization \citep{marx2024calibration} with a standard cross-entropy (XE) loss. Here, $n$ is the number of examples in the dataset, $d$ is the number of features $Z$, and $X$ is the covariate vector for which local calibration is evaluated. When the value of $p$ is less than 0.05, we reject the null hypothesis. } \label{tab:table2}\end{table*}
\section{Discussion} \label{discussion}

\noindent {\bf Distinction with related work.} There has been related work on local calibration \citep{marx2024calibration,luo2022local,park2021conditional}. These methods, like KLCE, are kernel-based and can be interchanged with minor modifications for testing and evaluating local calibration. One could construct other measures of local calibration and KLCE is not the only measure of local calibration. In principle, an appropriate decomposition of proper scoring rules can be used for evaluation of local calibration and as a measure of trustworthiness \citep{perez2022beyond}. There is an overlap between calibration literature and this work. The calibration literature has focused on a general trustworthiness framework, whereas this work develops an end-to-end trustworthiness framework for a specific context. \citet{vashistha2024u} argued against a general definition of trustworthiness and proposed that trustworthiness should be discussed and formalized in the context of a specific task. Following them, this work formalizes trustworthiness for inference tasks. We also note that KLCE and HSIC \citep{gretton2007kernel} are kernel-based hypothesis testing methods. HSIC measures independence between two random variables, whereas KLCE quantifies the extent of local miscalibration by integrating kernel methods with probabilistic classification metrics. Our hypothesis testing and diagnostic framework are specifically designed to address the calibration and trustworthiness in inference tasks, which differ from HSIC's original intent. There has also been work related to a competence-based framework for trustworthiness. Specifically, $\mathcal{U}$-trustworthiness employs a similar competence-based framework to formalize trustworthiness, with the key distinction that it focuses on utility maximization as the task of interest \citep{vashistha2024u}. As opposed to this work, to achieve $\mathcal{U}$-trustworthiness calibration or local calibration is unnessasary. Under certain assumptions, \cite{vashistha2024u} demonstrated that achieving $\mathcal{U}$-trustworthiness requires the model to produce rankings akin to those of the Bayes classifier. 

{\bf Limitations.} KLCE is sensitive to the choice of Kernel and how its hyperparameters are optimized, e.g., by minimizing Type-II error. We employ a greedy-based approach to set the kernel hyper-parameter, but future work should explore a better optimization algorithm. It is also necessary to emphasize that trustworthiness is necessary, but not sufficient, condition to achieve fairness in decision-making. There is no consensus on what constitutes an ``unbiased'' or ``fair'' algorithm. The existing definitions (except in trivial cases) cannot be satisfied simultaneously. However, the first step towards algorithmic fairness is to guarantee the trustworthiness of the information content of a predictive model. A user should not `blindly' apply it to their model without forward thinking about implications of an algorithm in a broader context. For an overview of algorithmic fairness literature, the reader may want to consult with the literature \citep{barocas2017fairness,chouldechova2018frontiers,corbett2018measure}.
\section{Conclusion}
This work introduces \ity\ as a framework for trustworthiness evaluation of probabilistic classifiers, when these models are utilized for inference tasks. We develop a hypothesis testing framework that employs a kernel-based test statistic, dubbed as KLCE, to assess local calibration. KLCE can be employed to (1) perform hypothesis testing, whether a probabilistic classifier is \ity, and (2) and quantify the level of untrustworthiness of a probabilistic classifier. Finally, we introduce a diagnostic method that allows the user to localize where in the feature space the model is biased and quantify its magnitude. The observation that multicalibration is a weaker condition than local calibration implies the necessity to invest in developing post-processing mitigation strategies. This work opens new directions towards local calibration techniques, which will complement the calibration and multicalibration literature.

\section*{Acknowledgements}
We acknowledge support from the National Science Foundation under Cooperative Agreement 2421782 and the Simons Foundation award MPS-AI-00010515. The authors would like to thank Dr. Stephen Walker, Bernardo F. Lopez and Giovanni Toto for their feedback. The authors also thank the reviewers for their helpful comments. 

\bibliography{references}
\bibliographystyle{apalike}

%%%%%%%%%%%%%%%%%%%%%%%%%%%%%%%%%%%%%%%%%%%%%%%%%%%%%%%%%%%%
\newpage

\newpage

\onecolumn

\section*{Checklist}

 \begin{enumerate}

 \item For all models and algorithms presented, check if you include:
 \begin{enumerate}
   \item A clear description of the mathematical setting, assumptions, algorithm, and/or model. 
   Answer: Yes, they are included both in the main text and the supplementary material. 
   \item An analysis of the properties and complexity (time, space, sample size) of any algorithm. 
   Answer. Yes, we have included an analysis of the properties such as Type-2 Error in the main text and Type-1 Error in the supplementary material.
   \item (Optional) Anonymized source code, with specification of all dependencies, including external libraries. [Yes/No/Not Applicable]
   Answer. The code is not made available currently but we will make it available after acceptance. 
 \end{enumerate}

 \item For any theoretical claim, check if you include:
 \begin{enumerate}
   \item Statements of the full set of assumptions of all theoretical results. 
   Answer. Yes, all assumptions are mentioned in the main text and supplemnetary material. 
   \item Complete proofs of all theoretical results. 
   Answer: Yes, all the proofy are discussed in the Supplementary Material.
   \item Clear explanations of any assumptions. Answer: Yes, all the assumptions are explained.    
 \end{enumerate}

 \item For all figures and tables that present empirical results, check if you include:
 \begin{enumerate}
   \item The code, data, and instructions needed to reproduce the main experimental results (either in the supplemental material or as a URL). Answer: Yes, the code is included in the zip file submitted as part of the supplementary material.
   \item All the training details (e.g., data splits, hyperparameters, how they were chosen). Answer: Yes, they are mentioned in the paper. 
    \item A clear definition of the specific measure or statistics and error bars (e.g., with respect to the random seed after running experiments multiple times). Answer: Yes, they are mentioned in the paper. 
    \item A description of the computing infrastructure used. (e.g., type of GPUs, internal cluster, or cloud provider). Answer: Not applicable.
 \end{enumerate}

 \item If you are using existing assets (e.g., code, data, models) or curating/releasing new assets, check if you include:
 \begin{enumerate}
   \item Citations of the creator If your work uses existing assets. Answer: Yes, it is included in the supplementary material. 
   \item The license information of the assets, if applicable. Answer: Yes, it is included in the supplementary material. 
   \item New assets either in the supplemental material or as a URL, if applicable. Answer: Yes, it is included in the supplementary material. 
   \item Information about consent from data providers/curators. Answer: Not Applicable
   \item Discussion of sensible content if applicable, e.g., personally identifiable information or offensive content. Answer: Not Applicable
 \end{enumerate}

 \item If you used crowdsourcing or conducted research with human subjects, check if you include:
 \begin{enumerate}
   \item The full text of instructions given to participants and screenshots. Answer: Not Applicable
   \item Descriptions of potential participant risks, with links to Institutional Review Board (IRB) approvals if applicable. Answer: Not Applicable
   \item The estimated hourly wage paid to participants and the total amount spent on participant compensation. Answer: Not Applicable
 \end{enumerate}

 \end{enumerate}

\appendix

\section{Supplementary Material}
\subsection{Bootstrap Algorithm}

The statistical significance ($p$-value) is defined as the probability of the null distribution exceeding the estimated KLCE for a proposed model. The $p$-value can be computed as:  
\begin{equation} \label{eq:p-value-computations}
    p = {\rm Pr}\left(\widehat{{\rm KLCE}}^2_{\rm null}[k, y, \whf] > \widehat{{\rm KLCE}}^2_{\rm data}[k, y, \whf]\right).
\end{equation}
Algorithm \ref{alg:bootstrap} shows a simulation-based bootstrap algorithm used to construct the distribution under the null hypothesis and estimate Equation \eqref{eq:p-value-computations} numerically.

\begin{algorithm}[ht!]
\caption{Simulation-based bootstrap algorithm for estimating the null distribution and p-value}
\label{alg:bootstrap}
\begin{algorithmic}[1]

\Statex \textbf{Input}: 
  \(\mathbf{x}\) (test samples), 
  \(\widehat{f}\) (classifier predictions),
  \(\mathbf{y}\) (actual observations), 
  \(k(\cdot,\cdot)\), \(l(\cdot,\cdot)\) (kernel functions),
  \(N_b\) (number of bootstraps)

\Statex \textbf{Output}:
  \(\widehat{\mathrm{KLCE}}^2_{\rm data}\) (KLCE\textsuperscript{2} on observed data), 
  \(\{\widehat{\mathrm{KLCE}}^2_{\rm null}[i]\}\) (null distribution),
  \(p\)-value

\vspace{0.3em}
\State Set hyperparameters for \(k(\cdot,\cdot)\) and \(l(\cdot,\cdot)\).
\State Compute residuals: \(\mathbf{e} \leftarrow \mathbf{y} - \widehat{f}\).
\State Compute kernel matrix: 
  $K \leftarrow \mathbf{e}^{\top}\,k(\widehat{f},\widehat{f})\,l(\mathbf{x},\mathbf{x})\,\mathbf{e}$.
\State Compute 
$\widehat{\mathrm{KLCE}}^2_{\rm data} \leftarrow \frac{\sum(K) - \sum(\text{diag}(K))}{n(n-1)}$.
\State Initialize an empty list for \(\widehat{\mathrm{KLCE}}^2_{\rm null}\).

\For{\(i = 1 \to N_b\)}
  \State Draw \(n\) i.i.d.\ samples \(\mathbf{e_b}\) from \(\mathbf{e}\) (with replacement).
  \State Compute 
    $
      K_b \leftarrow \mathbf{e_b}^{\top}\,k(\widehat{f},\widehat{f})\,l(\mathbf{x},\mathbf{x})\,\mathbf{e_b}.
    $
  \State Calculate $\widehat{\mathrm{KLCE}}^2_{\rm null}[i] \leftarrow \frac{\sum(K_b) - \sum(\text{diag}(K_b))}{n(n-1)}.
    $
\EndFor

\State \(\displaystyle p\text{-value} \leftarrow \frac{\#\{\widehat{\mathrm{KLCE}}^2_{\rm null}[i] > \widehat{\mathrm{KLCE}}^2_{\rm data}\}}{N_b}.\)

\State \textbf{Return} \(\widehat{\mathrm{KLCE}}^2_{\rm data},\; \{\widehat{\mathrm{KLCE}}^2_{\rm null}[i]\},\; p\text{-value}.\)

\end{algorithmic}
\end{algorithm}

\subsection{Type-I Error} 
Here, we show that Type-I error is determined by the null hypothesis rejection threshold and is independent of kernel choice, kernel parameters, and sample size.

\begin{figure}[H]
\centering
\includegraphics[width=0.45\textwidth]{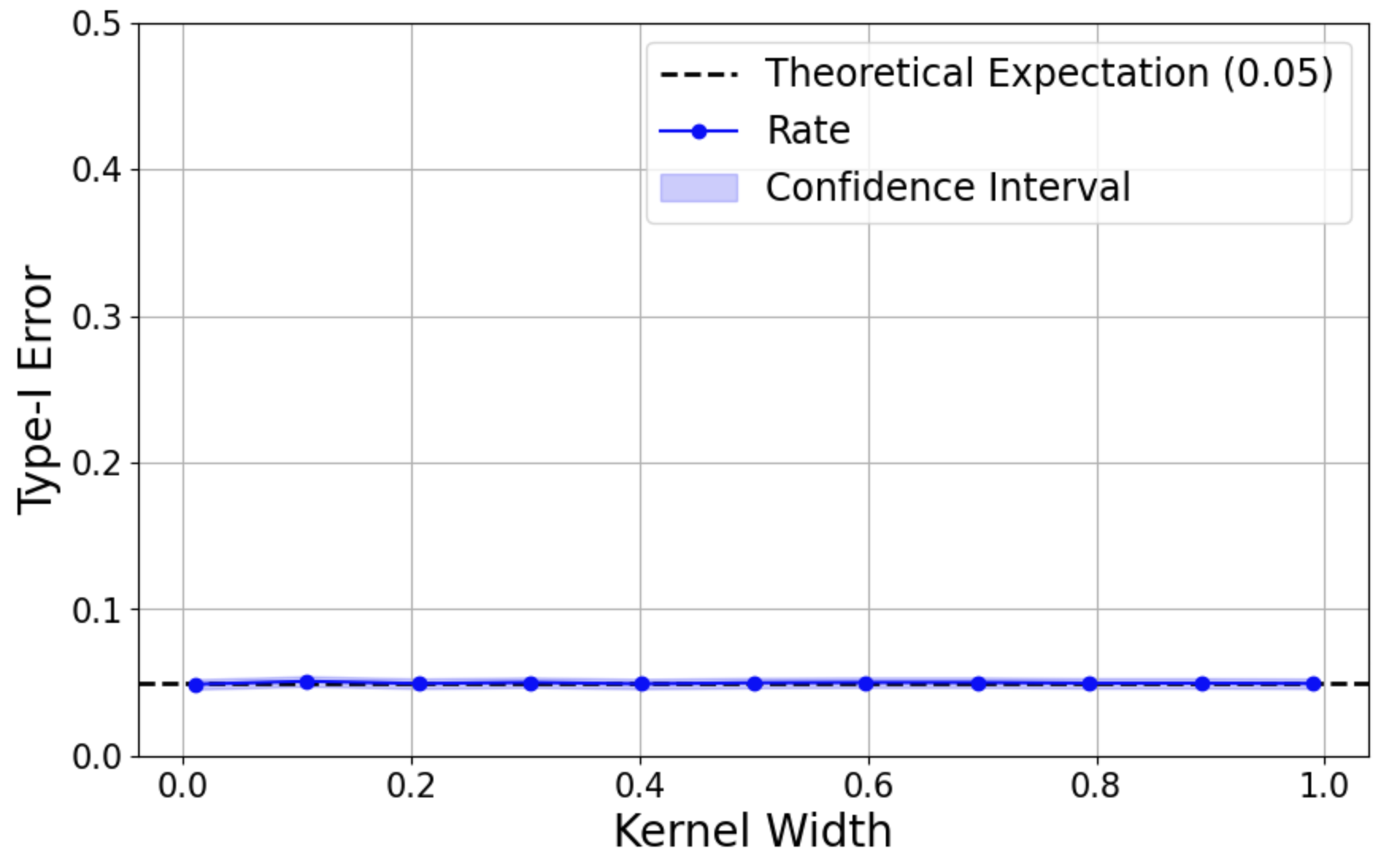}
\includegraphics[width=0.45\textwidth]{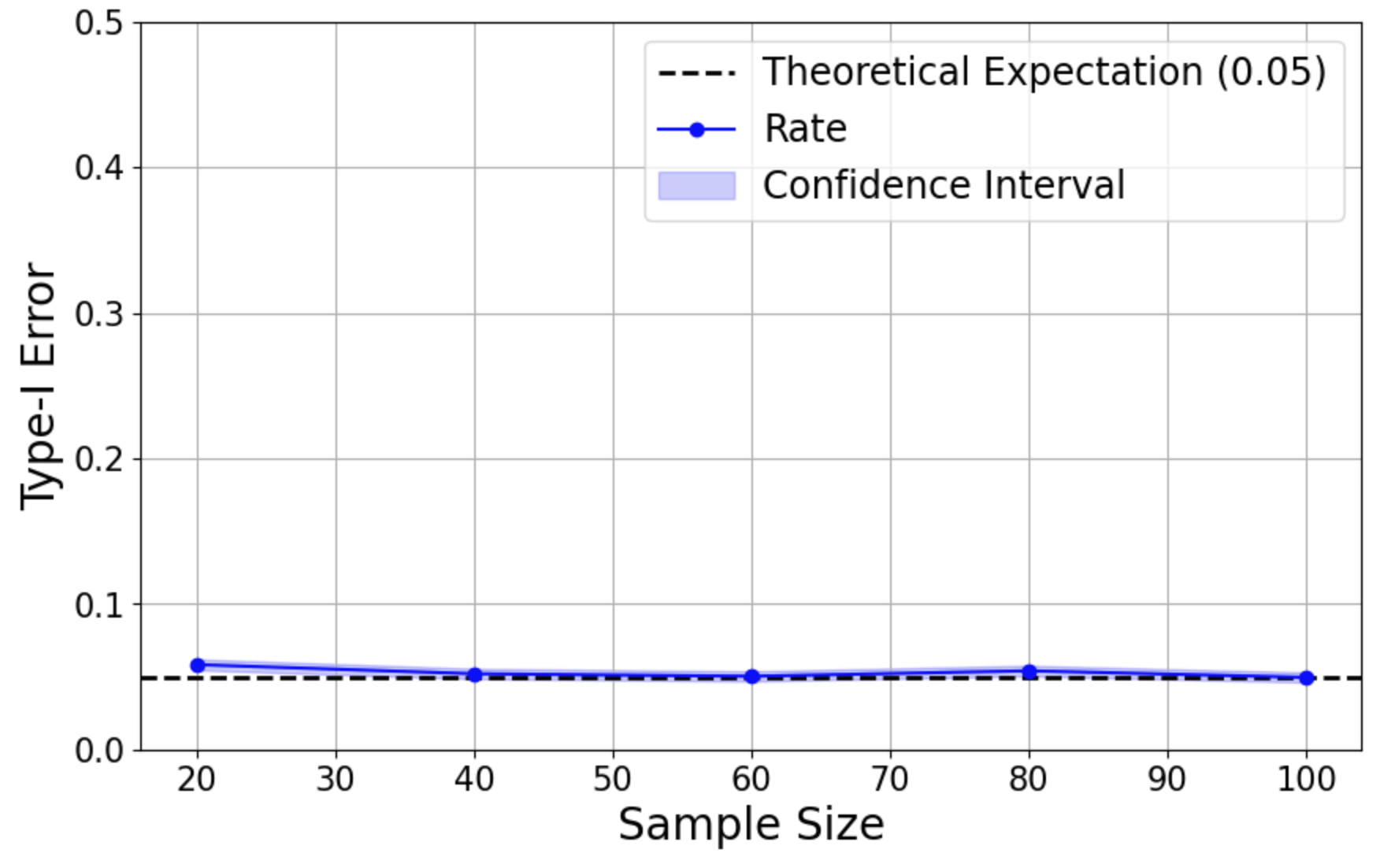}
\vspace{-2mm}
\caption{Type-I Error as a function of kernel width and sample size.}
\end{figure}

\subsection{Data Sets}

\subsubsection{COMPAS}

The COMPAS dataset\footnote{Link: \url{https://github.com/propublica/compas-analysis/blob/master/compas.db}} is derived from a 2016 investigation by ProPublica into the COMPAS (Correctional Offender Management Profiling for Alternative Sanctions) scoring system, which is used by U.S. courts to assess the risk of recidivism among defendants \cite{Angwin2016MachineBias,Propublica2016CompasAnalysis}. This dataset includes information on 2,714 defendants from Broward County, Florida, collected during 2013-2014, and features variables such as age, race, criminal history, COMPAS recidivism scores, and whether the individual reoffended within two years. The dataset has a CC-BY 4.0 license.  

\subsubsection{Homeownership}

This study utilizes the homeownership dataset from the 2019 American Housing Survey \citep{farahi2024analyzing}, %\footnote{\url{https://raw.githubusercontent.com/afarahi/Scientific-Machine-Learning/main/AHS\_2019\_Cleaned.CSV}}
a detailed and comprehensive national survey on housing in the U.S., conducted by the Census Bureau and sponsored by the Department of Housing and Urban Development. The dataset captures a variety of household characteristics, including income, marital status, education level, and more. It also includes division, metropolitan status, race, and gender variables, detailed in Table~\ref{tab:vars}. The work specifically examines households with incomes ranging from \$10,000 to \$500,000. After excluding non-respondents and households outside this income range, the dataset includes 48,660 households. For the modeling process, a log-income transformation has been applied, with no other data modifications. The dataset has a CC-BY 4.0 license.   
\begin{table}[H]
\centering
\caption{Independent and dependent variables.} \label{tab:vars}
\begin{tabular}{|l|l|l|l|}
\hline
{\bf Attribute Name} & {\bf Definition}  &  {\bf Type } & {\bf Values}  \\ \hline 
BLACK          & One if at least one member of the     & Binary  & \{0, 1\}        \\
              &  household is African-American, 0 otherwise &      &         \\\hline
HHMAR          & Marital status                                                         & Categorical & -- \\ \hline
HINCP          & Household income                                                       & Numeric     & {[}\$10k, \$500k{]}  \\ \hline
HHGRAD         & Educational attainment of householder                                  & Categorical     & -- \\ \hline
HHAGE          & Age of householder                                                      & Numeric     & {[}15, 85{]}    \\ \hline
DIVISION       & Division                                                               & Categorical & -- \\ \hline
HHCITSHP       & Citizenship of householder                                              & Categorical & -- \\ \hline
NUMPEOPLE      & Count of households                                                    & Numeric     & {[}1, 18{]}     \\ \hline
HHSEX          & Sex of householder (Male=1, Female=2)                                       & Binary      & \{1, 2\}     \\ \hline
 METRO          & One if in a metropolitan area, 0 otherwise.                            & Binary      & \{0, 1\}         \\\hline 
OWNER          & One if owned or being bought by someone  & Binary      & \{0, 1\} \\       
          &  in the household, 0 otherwise. &      &  \\\hline
\end{tabular}
\end{table}

\subsection{Local Calibration Bias}

\begin{figure*}[ht]
    \centering 
    \includegraphics[width=1\textwidth]{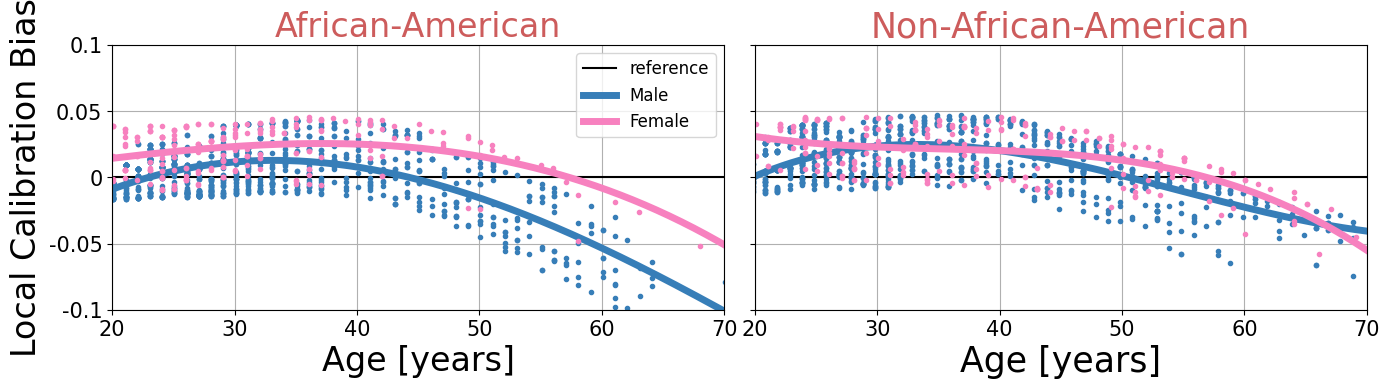} 
    \vspace{-9mm}
    \caption{Local calibration bias estimated for a subset of defendants. The lines are third-order polynomial fits to the data that show the average calibration bias per group as a function of age.} \vspace{-3mm}
    \label{fig:results_bias}
\end{figure*}

Figure \ref{fig:results_bias} shows the estimated local bias for a subset of defendants in the test sample, selected based on their gender (blue vs. pink points) and race (left vs. right panel). The lines are third-order polynomial fits to the data stratified based on race and age and show the average calibration bias per group as a function of the defendant’s age. For example, the pink line in the left panel shows the aggregated bias for all African-American females within a particular age group. This figure suggests that the calibration bias is a function of race, gender, and age. The prediction model systematically underestimates the risk for female (pink line) and male (blue line) African-American (left panel) and non- African-American (right panel) defendants of age 30 to 40. There is a systemic prediction bias difference between male and female African-American defendants. There is also a non-linear trend with respect to the age of defendants for all four groups. The risk prediction bias cannot be studied for each feature independent of others, but the model can capture a non-linear, group-dependent bias of a predictive model.

\subsection{Kernel Choice}

For all the experiments in the paper, we assume a RBF kernel for both $f(.)$ and $X$. Our theoritical results hold for all universal kernels $k(.)$ and $l(.)$.

\subsection{Proof of Lemmas, Propositions, and Theorems}

For completeness and readability, we include all the definitions, lemmas, proofs, and theorems mentioned in the main text.

\subsubsection{Notations}
We consider a probabilistic binary classification problem where $Y \in \{0, 1\}$ is the class label and $Z\in \calZ$ and $X \in \calX$ are, respectively, $d_{\bfz}$-dimensional and $d_{\bfx}$-dimensional instances of two, potentially overlapping, feature vectors. $Z$ is the classifier input, and $X \in \calX$ is an instance of a secondary feature vector for which we want to evaluate the model's trustworthiness. A probabilistic binary classifier, denoted with $\whf : \calZ \rightarrow \Delta$, outputs class probabilities $\whf(Z)$ of $p_{Y=1|Z}(y=1\mid z)$ for all $z \in \mathcal{Z}$. $\Delta$ denotes the probability simplex $\Delta := \{\whf(z) \in \mathbb{R}_{\geq 0}: \norm{\whf(z)}_1 = 1 \}$. 

Similarly, we denote the conditional distribution of $Y = 1$ given $\whf(Z)$ by $p_{Y=1|\whf(Z)}(y=1 \mid \whf(z))$ and conditional distribution of $Y=1$ given $\whf(Z)$, $X$ by $p(y=1 \mid \whf(z),x)$ for all $\whf(z) \in \Delta$ and $x \in \mathcal{X}$. Now, we can define a function $\delta:\Delta\rightarrow\mathbb{R}$, given by
$$
\delta(\whf(z)):= p_{Y=1 \mid \whf(Z)}(y=1|\whf(z))-\whf(z)
$$
for formulating calibration in strong way. A model $\whf$ is calibrated in the strong sense if $p_{Y = 1 \mid \whf(Z)}(y=1 \mid \whf(z))=\whf(z)$ almost surely for all $\whf(z) \in \Delta$, or equivalently if $\delta(\whf(z))=0$ almost surely for all $\whf(z) \in \Delta$. For ease of notation, we use $\delta$ in remainder of the text instead of $\delta(\whf(z))$. 

Similarly, to formulate local calibration, we may define another function $\delta_c: \Delta \rightarrow \mathcal{R}$ where 
$$
\delta_c(\whf(z),x):=p_{Y = 1\mid \whf(Z),X}(y=1 \mid \whf(z),x)-\whf(z)
$$
Now, a model $\whf$ is locally calibrated in the strong sense if $p_{Y =1  \mid \whf(Z), X}(y = 1 \mid \whf(z), x)=\whf(z)$ almost surely for all $\whf(z) \in \Delta$ and $x \in \mathcal{X}$, or equivalently if $\delta_c(\whf(z),x)=0$ almost surely for all $\whf(z) \in \Delta$ and $x \in \mathcal{X}$. For ease of notation, we use $\delta_c$ in the remainder of the text instead of $\delta_c(\whf(z),x)$. 

\subsubsection{Calibration Error}

\begin{definition}(Calibration error). Let $\mathcal{H}$ be a class of functions h: $ \Delta \rightarrow \mathbb{R}$ . Then the calibration error $\mathrm{CE}$ of model $\whf$ with respect to class $\mathcal{H}$ is
$$
\mathrm{CE}[\mathcal{H}, \whf]:= \sup _{h \in \mathcal{H}} \mathbb{E}[\delta(\whf(Z)) \times h(\whf(Z))] = \sup _{h \in \mathcal{H}} \mathbb{E}\left[\langle \delta, h(\whf(Z))\rangle_{\mathbb{R}}\right]
$$
where $\mathcal{H}$ is a unit ball in a RKHS $\mathcal{F}$.
\end{definition}

\begin{definition}(Local Calibration error).  Let $\mathcal{H}$ be a class of functions h: $ \Delta \rightarrow \mathbb{R}$. Let $\mathcal{G}$ be a class of functions g: $ \mathcal{X} \rightarrow \mathbb{R}$. Then the local calibration error $\mathrm{LCE}$ of model $\whf$ with respect to class $\mathcal{H}$ and class $\mathcal{G}$ is
$$
\mathrm{LCE}[\mathcal{H}, \mathcal{G}, \whf]:= \sup _{h \in \mathcal{H}, g \in \mathcal{G}} \mathbb{E}[\delta_{c}(\whf(Z),X) \times h(\whf(Z))g(X)] =  \sup _{h \in \mathcal{H}, g \in \mathcal{G}} \mathbb{E}\left[\langle \delta_c, h(\whf(Z))g(X)\rangle_{\mathbb{R}}\right]
$$
where $\mathcal{H}$ is a unit ball in a RKHS $\mathcal{F}$ and $\mathcal{G}$ is a unit ball in a RKHS $\mathcal{K}$.
\end{definition}

Consider the outer product  $\mathcal{M}$ = $\mathcal{H} \otimes \mathcal{G}$ where for each $h \in \mathcal{H}$ and $g \in \mathcal{G}$, we define a function $h \otimes g$ such that $(h \otimes g)(\whf(Z),X) = h(\whf(Z))g(X)$. This forms a new class of functions which belong to the RKHS $ \mathcal{P} = \mathcal{F} \otimes \mathcal{K}$ .

Let $k: \Delta \times \Delta \rightarrow \mathbb{R}$ be the kernel of RKHS $\mathcal{F}$ and $l: \mathcal{X} \times \mathcal{X} \rightarrow \mathbb{R}$ be the kernel of RKHS $\mathcal{K}$. Then kernel $p$ of $\mathcal{P} = \mathcal{F} \otimes \mathcal{K}$ is tensor product of kernels k and l of $\mathcal{F}$ and $\mathcal{K}$ respectively, defined as
$$
p((\whf(Z),X),(\whf(Z'),X')) = k(\whf(Z),\whf(Z'))l(X,X')
$$

Now we reformulate the LCE in terms of $\mathcal{M}$, 
$$
\mathrm{LCE}[\mathcal{H}, \mathcal{G}, \whf]:= \sup _{m \in \mathcal{M}} \mathbb{E}[\delta_{c} \times m(\whf(Z),X)]
$$
where $m(\whf(Z),X) = h(\whf(Z))g(X)$ for $h \in \mathcal{H}$ and $g \in \mathcal{G}$.
This reformulation allows us to connect calibration with local calibration and
 use results from \cite{Widmann:2019calibration}, which dealt with the problem of calibration. 

\begin{lemma} \label{lemma-lceis0}
 $\mathrm{LCE}[\mathcal{H}, \mathcal{G}, \whf] = 0$ if and only if model $\whf$ is locally calibrated in the strong sense.    
\end{lemma}

\begin{proof}
If model $\whf(.)$ is locally calibrated in the strong sense, then $\delta_c=0$ almost surely. Hence for all $h \in \mathcal{H}$ and $g \in \mathcal{G}$, we have that $\mathbb{E}[\delta_{c} \times h(\whf(Z))g(X)]=0$, which implies $\mathrm{LCE}[\mathcal{H}, \mathcal{G}, \whf] = 0$.
\\ \\
We prove the converse to show that when $\whf(.)$ is not locally calibrated in the strong sense, then $\mathrm{LCE}[\mathcal{H}, \mathcal{G}, \whf] \neq 0$.
\\
When $\delta_c \neq 0$ almost surely, then there exists a $\whf(z^{*})= \alpha^{*}$ \& $X=x^{*}$, where $\delta_c(\alpha^{*},x^{*}) = \gamma$ where  
$\gamma \neq 0 $. We then consider a $m \in \mathcal{M}$ to be a function such that it takes its maximum value around $(\alpha^{*},x^{*})$  and is zero everywhere else. Then, we would have that $\mathbb{E}[\delta_{c} \times m(\whf(Z),X)] \neq 0$
\end{proof}

\begin{lemma}(Existence and uniqueness of embedding)\label{embedding-lemma}
Let $\mathcal{P}$ be a RKHS with kernel $p: (\Delta, \mathcal{X}) \times (\Delta, \mathcal{X}) \rightarrow \mathbb{R}$, and assume that $p(\cdot, t) u$ is measurable for all $t \in (\Delta, \mathcal{X})$ and $u \in \mathbb{R}$, and $\|p\|<\infty$.
Then there exists a unique embedding $\mu \in \mathcal{P}$ such that for all $m \in \mathcal{P}$
$$
\mathbb{E}\left[\langle\delta_c, m(\whf(Z),X)\rangle_{\mathbb{R}}\right]=\left\langle m, \mu \right\rangle_{\mathcal{P}} .
$$
The embedding $\mu$ satisfies for all $t \in (\Delta, \mathcal{X})$ and $y \in \mathbb{R}$
$$
\left\langle y, \mu(t)\right\rangle_{\mathbb{R}}=\mathbb{E}\left[\langle\delta_c, p((\whf(Z),X), t) y\rangle_{\mathbb{R}}\right]
$$  
\end{lemma}

\begin{theorem}(Explicit formulation) \label{th-formula}
Let $\mathcal{P}$ be a RKHS with kernel $p: (\Delta, \mathcal{X}) \times (\Delta, \mathcal{X}) \rightarrow \mathbb{R}$, and assume that $p(\cdot, t) u$ is measurable for all $t \in (\Delta, \mathcal{X})$ and $u \in \mathbb{R}$, and $\|p\|<\infty$.
Then
$$
\mathrm{LCE}[\mathcal{M}, \whf]=\left\|\mu\right\|_{\mathcal{P}},
$$
where $\mu$ is the embedding defined in Lemma \ref{embedding-lemma}. Moreover,
$$
\mathrm{KLCE}[k,l,\whf]^2:=\operatorname{LCE}^2[\mathcal{M}, \whf]=\mathbb{E}[(Y-\whf(Z))k(\whf(Z),\whf(Z'))l(X,X')(Y'-\whf(Z'))] ,
$$
where $\left(X^{\prime}, Z^{\prime}, Y^{\prime}\right)$ is an independent copy of $(X, Z, Y)$.    
\end{theorem} 
\begin{proof}
Let $\mathcal{M}$ be a unit ball in RKHS $\mathcal{P}$. We now simplify LCE using kernels
$$
\begin{aligned}
   \mathrm{LCE}[\mathcal{M}, \whf]&= \sup _{m \in \mathcal{M}} \mathbb{E}[\delta_{c} m(\whf(Z),X)]
   \\ &= \sup _{m \in \mathcal{M}} \langle m, \mu \rangle  \\
   &= \norm{\mu}_{\mathcal{P}}
\end{aligned}
$$
Thus using Lemma \ref{embedding-lemma} we obtain,
$$
\begin{aligned}
    \mathrm{KLCE}^{2}[l,k,\whf] = \mathrm{LCE}[\mathcal{M}, \whf]^2 &= \langle \mu, \mu \rangle \\
    &= \mathbb{E}[\langle \delta_c, \mu(\whf(Z), X) \rangle] \\
    &=  \mathbb{E}[\mathbb{E} \langle \delta_c', p((\whf(Z), X), (\whf(Z'), X'))\delta_c| \whf(Z),X \rangle] \\
    &= \mathbb{E}[\langle \delta_c, p((\whf(Z), X), (\whf(Z'), X'))\delta_c' \rangle] \\
    &= \mathbb{E}[\langle \delta_c, k(\whf(Z),\whf(Z'))l(X,X')\delta_c' \rangle] \\
    &= \mathbb{E}[\delta_c k(\whf(Z),\whf(Z'))l(X,X')\delta_c']
\end{aligned}
$$
By rewriting 
$$
\delta_c = \mathbb{E}[Y|\whf(Z),X] - \whf(Z) = \mathbb{E}[Y-\whf(Z)|\whf(Z),X]
$$
and $\delta_c^{'}$ in the same way, we get
$$
\mathrm{KLCE}^{2}[l,k,\whf] = \mathbb{E}[(Y-\whf(Z))k(\whf(Z),\whf(Z'))l(X,X')(Y'-\whf(Z'))] 
$$
\end{proof}

\subsubsection{Estimators}
The estimator of $\mathrm{KLCE}^2[l,k, \whf]$ is
$$
\widehat{\mathrm{KLCE}^2}[k, l, \{x, y, z\}, \whf]:= \frac{1}{n(n-1)}  \sum_{j \neq i}^n [(y_i - \whf(z_i))k(\whf(z_i),\whf(z_j))l(x_i,x_j)(y_j-\whf(z_j))]
$$
where $y_{i/j},x_{i/j}$ and $z_{i/j}$ are instances of $(Y_{i/j}, X_{i/j}, Z_{i/j})$. 

\begin{theorem} \label{theorem-unbiased}
$\widehat{\mathrm{KLCE}^2}[k, l, \whf]$is an unbiased estimator of $\mathrm{KLCE}^2[k, l, \whf]$
\end{theorem}
 
\begin{proof} 
We know that,
$$
\mathrm{KLCE}^{2}[k, l, \whf] = \mathbb{E}[(Y-\whf(Z))k(\whf(Z),\whf(Z'))l(X,X')(Y'-\whf(Z'))] 
$$
where $\left(X^{\prime}, Z^{\prime} ,Y^{\prime}\right)$ is an independent copy of $(X, Z, Y)$. Since $\left(x_i, z_i, y_i\right)$ are i.i.d., we have
$$
\begin{aligned}
\mathbb{E}\left[\widehat{\mathrm{KLCE}^2}[k, l, , \{x, y, z\}, \whf]\right] & =\frac{1}{n(n-1)}  \sum_{j \neq i}^n \mathbb{E}[(y_i - \whf(z_i))k(\whf(z_i),\whf(z_j))l(x_i,x_j)(y_j-\whf(z_j))]\\
& = \mathbb{E}[(Y-\whf(Z))k(\whf(Z),\whf(Z'))l(X,X')(Y'-\whf(Z'))]\\
& = \mathrm{KLCE}^{2}[k, l, \whf]
\end{aligned}
$$
which shows that $\widehat{\mathrm{KLCE}^2}[k, l, \{x, y, z\}, \whf]$ is an unbiased estimator of $\mathrm{KLCE}^{2}[k, l, \whf]$.
\end{proof}

\subsubsection{Calibration Tests}

\begin{lemma} \label{lemma-condition}
Let $p: (\Delta, \mathcal{X}) \times (\Delta, \mathcal{X}) \rightarrow \mathbb{R}$ be a kernel, and assume that $P_{\alpha ; \beta}:=$ $\sup _{s, t \in (\Delta \times \mathcal{X})}\|p(s, t)\|_{\alpha ; \beta}<\infty$ for some $1 \leq \alpha, \beta < \infty$. Then
$$
\sup _{z, z^{\prime} \in \mathcal{Z}, x, x^{\prime} \in \mathcal{X}}\left|\left\langle\delta_c, p\left( (f\left(z\right),x), (f\left(z'\right),x')\right) \delta_c^{\prime})\right\rangle_{\mathbb{R}}\right| \leq 2^{1+1 / \alpha-1 / \beta} P_{\alpha ; \beta}=: B_{\alpha ; \beta} .
$$  
\end{lemma}

\begin{proof}
The proof follows that of Lemma H.1 in Widmann et al. (2019).
\end{proof}

\begin{theorem} \label{th: test-proofs}
Let $p: (\Delta, \mathcal{X}) \times (\Delta, \mathcal{X}) \rightarrow \mathbb{R}$ be a kernel, and assume that $p(\cdot, t) u$ is measurable for all $t \in \Delta$ and $u \in \mathbb{R}$, and $P_{\alpha ; \beta}:=\sup _{s, t \in \Delta}\|p(s, t)\|_{\alpha ; \beta}<\infty$ for some $1 \leq \alpha, \beta < \infty$. Then for all $\epsilon>0$
$$
\mathbb{P}\left[|\widehat{\mathrm{KLCE}^2}[k, l, \{x, y, z\}, \whf]-\operatorname{KLCE}^2[k, l, f]| \geq \epsilon\right] \leq 2\exp \left(-\frac{\epsilon^2 n}{2 B_{\alpha ; \beta}^2}\right)
$$
\end{theorem}

\begin{proof}
By Theorem \ref{theorem-unbiased}, we know that $\mathbb{E}(\widehat{\mathrm{KLCE}^2}[k, l, \{x, y, z\}, \whf]) = \mathrm{KLCE}^{2}[k, l,\whf]$. Now, using Theorem \ref{th-formula}, we rewrite $\widehat{\mathrm{KLCE}^2}[k, l, \{x, y, z\}, \whf]$ as
$$
\widehat{\mathrm{KLCE}^2}[k, l, \{x, y, z\}, \whf]= \frac{1}{n(n-1)} \sum_{i \neq j}^{n} \langle \delta_c, p\left( (f\left(z\right),x), (f\left(z'\right),x')\right) \delta_c^{\prime} \rangle
$$
We also have from Lemma \ref{lemma-condition} that 
$$
\sup _{z, z^{\prime} \in \mathcal{Z}, x, x^{\prime} \in \mathcal{X}}\left|\left\langle\delta_c, p\left( (f\left(z\right),x), (f\left(z'\right),x')\right) \delta_c^{\prime})\right\rangle_{\mathbb{R}}\right| \leq  B_{\alpha ; \beta} .
$$ 
Thus, by Hoeffding's inequality for all $ \epsilon > 0$, we have
$$
\mathbb{P}\left[|\widehat{\mathrm{KLCE}^2}[k, l, \{x, y, z\}, \whf]-\operatorname{KLCE}^2[k, l, f]| \geq \epsilon\right] \leq 2\exp \left(-\frac{\epsilon^2 n}{2 B_{\alpha ; \beta}^2}\right)
$$
\end{proof}

\subsubsection{Local Calibration Bias}
\begin{proposition}
The supremum function, referred to as local calibration bias (LCB), has a closed-form solution \cite{Gretton:2012kernel}. The solution yields
\begin{align} \label{eq:EWF}
    &{\rm LCB}[k, l,  \{x^{\prime}, y^{\prime}, z^{\prime}\},\whf] :=\bbE \left[ \delta_c \times k(\whf(z^{\prime}), \whf(z))\,l(x, x^{\prime})) \right]. \nonumber
\end{align}

\begin{proof}
The local calibration bias is defined to be the RKHS function that maximizes the supremum in Equation \ref{eq: lce}.

Now, we know that LCE can be rewritten as
$$
\mathrm{LCE}[\mathcal{H}, \mathcal{G}, \whf]:= \sup _{m \in \mathcal{M}} \mathbb{E}[\delta_{c} \times m(\whf(Z),X)]
$$
where $m(\whf(Z),X) = h(\whf(Z))g(X)$ for $h \in \mathcal{H}$ and $g \in \mathcal{G}$. Here, we want to find a function $m$ that maxmizes the supremum. 

Let $\mathcal{M}$ be a unit ball in RKHS $\mathcal{P}$. We now simplify LCE using kernels as follows:
\begin{equation}
   \mathrm{LCE}[\mathcal{M}, \whf]= \sup _{m \in \mathcal{M}} \mathbb{E}[\delta_{c} m(\whf(Z),X)] = \sup _{m \in \mathcal{M}} \langle m, \mu \rangle  
\end{equation}
Now, using Lemma \ref{embedding-lemma}, we  notice the function $m(.)$ that maximizes the expression above is $m(t) = \mu(t) = \mathbb{E}\left[\langle\delta_c, p((\whf(Z),X), t) \rangle_{\mathbb{R}}\right]$ where, $t \in (\Delta,\mathcal{X})$.

Thus, 
$${\rm LCB}[k, l, \whf, \{x^{\prime}, y^{\prime}, z^{\prime}\}] = \mathbb{E}\left[\delta_c \times p((\whf(z),x), (z^{\prime},x^{\prime}))\right] = \bbE \left[ \delta_c \times k(\whf(z^{\prime}), \whf(z))\,l(x, x^{\prime})) \right]. \nonumber$$

\end{proof}
    
\end{proposition}
    
\end{document}